\documentclass[runningheads]{llncs}
\usepackage{graphicx}
\usepackage{amsmath,amssymb} 
\usepackage{ruler}
\usepackage{color}
\usepackage{multicol}
\begin{document}
\pagestyle{headings}
\mainmatter
\def\ACCV18SubNumber{***}  
\title{Partial Person Re-identification with Alignment and Hallucination}
\author{Sara Iodice \thanks{Principal Contact: s.iodice16@imperial.ac.uk} and Krystian Mikolajczyk \thanks{k.mikolajczyk@imperial.ac.uk}}
\institute{Imperial College London, UK}
\maketitle
\begin{abstract}
Partial person re-identification involves matching pedestrian frames where only a part of a body is visible in corresponding images. This reflects practical CCTV surveillance scenario, where full person views are often not available. Missing body parts make the comparison very challenging due to significant misalignment and varying scale of the views.  
We propose Partial Matching Net (PMN) that detects body joints, aligns partial views and hallucinates the missing parts based on the information present in the frame and a learned model of a person.  The aligned and reconstructed views are then combined into a joint representation and used for matching images. We evaluate our approach and compare to other methods on three different datasets, demonstrating significant improvements.
\end{abstract}
\section{Introduction}
\label{sec:intro}
Research in person re-identification (re-ID) has advanced with CNN base methods and their performance on academic datasets almost saturated. However, these datasets do not reflect well a typical real application scenario. 
Images in widely used benchmarks \cite{li2014deepreid,zheng2015scalable,gray2008viewpoint} have similar viewing angle, contain most of the human body, and are well aligned.  Many of the state of the art methods assume full person view in both probe and the gallery.
In real scenarios, a picture of a person often contains only a partial view, e.g. in crowded scenes.
Partial person re-ID aims at recognizing or matching identities of people from frames containing only partial views of human bodies, that are acquired by different and non overlapping cameras.
\begin{figure}[h]
\centering
{\includegraphics[height=2.6cm]{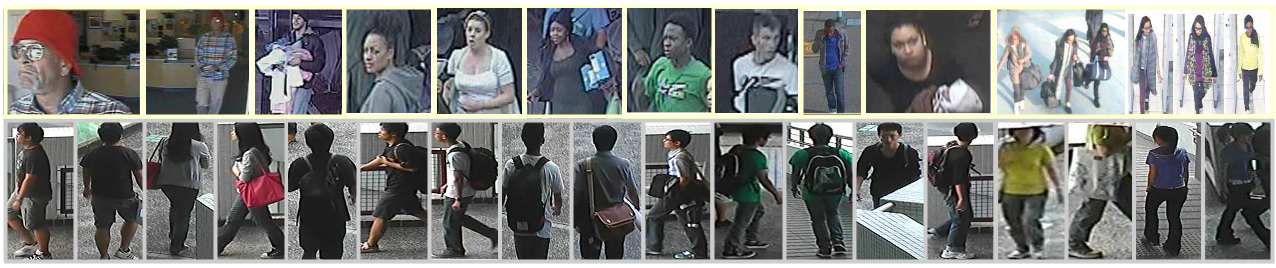}}
\caption{Example images or real police cases (top) and samples from academic datasets (bottom). Images in standard benchmarks such as CUHK03, have similar viewing angle, are better aligned and contain most of the human body.}
\label{fig:cover}
\end{figure}
For example, in figure \ref{fig:cover} real cases of people wanted by the police (top row) can be compared to samples from standard datasets such as CUHK03 (bottom row). Even with some body parts missing due to applied automatic person detector the views are still significantly more complete than the real cases.
In addition, real images are not aligned making the re-ID even more challenging.
\newline
Despite partial person re-ID being of high relevance to practical applications, it is still little addressed in the literature. Very few datasets were proposed for this task, such as simulated partial REID and partial i-LIDS \cite{zheng2015partial}, having a limited number of pedestrians/frames, insufficient for training modern CNN based models. In addition, the gallery set consists of full body views making it still distant from the real scenarios. 
\newline
To bridge the gap between the simulated and real re-ID task, we generate new and more challenging datasets from the widely used CUHK03 \cite{li2014deepreid}.
Furthermore, to address the partial Re-ID task, we propose an approach based on two strategies, alignment, and hallucination, motivated by the following observations. Firstly, alignment is a crucial process in human perception facilitating direct comparison between corresponding parts.
Secondly, in human vision, missing content that is due to the eye blind spot is hallucinated in the brain, based on the context and prior knowledge. Although the hallucinated part may not correspond to the real view, it improves our ability to recognize objects.
Similarly, we conjecture that alignment and hallucination may boost the recognition process in case of partial re-ID, by reconstructing missing coarse structures of the human body. 
From a practical perspective, CNN filters are adapted to spatial arrangement of parts and do not cope well with significant misalignment or large parts missing in some views, therefore we expect that providing full and aligned views should lead to improvements in partial person re-ID.
We propose a novel architecture which is jointly optimized with real and hallucinated samples.
To the best of our knowledge, this is the only work studying the impact of hallucination on human re-ID task.
This idea has been exploited in face recognition \cite{xu2013face} jointly optimized with real and hallucination face examples. 
In summary, we make the following contributions: 
a) introduce an approach to partial re-ID that combines body joint detection, alignment, and hallucination;
b) demonstrate the impact of image alignment in case of partial views;
c) generate  dataset \textit{CropCUHK03} which reflects partial re-ID  scenario better than the existing datasets;
e) validate the approach in an extensive evaluation and comparison to several methods on three different datasets;
f) show that the proposed approach leads to significant improvements in partial person re-ID.
\section{Related work}
\textbf{Partial person re-ID.}
Whilst many works have focused on the traditional task of full person re-ID,  the more challenging task of partial person re-ID has received little attention.
Partial re-ID was recently introduced in \cite{zheng2015partial} where partial observations of the query images were used, however, the gallery images contained full persons.  Ambiguity-Sensitive Matching Classifier (AMC)\cite{zheng2015partial} was also proposed with two matching approaches i.e. a local-to-local based on matching small patches, and global-to-local with a sliding window search where the partial observation served as a template.  The proposed approach was validated on small datasets i.e. Partial Reid, i-LIDS and Caviar, and may not scale well to real scenarios due to the quadratic cost.
More recently,   Deep Pixel Reconstruction (DPR) \cite{he2018deep} reconstructs missing channels of the query feature maps from full observations in the gallery maps. 
\newline
Unlike these methods, we align and reconstruct missing body parts by using contextual and training samples only, without full body views in the gallery set. 
\newline
\textbf{Person re-ID.}
A number of CNN based solutions have been proposed with diverse complexity of feature learning, or metric learning.
Several works introduce specific layers and new components to learn strongest features against variations in people appearance across multiple cameras and misalignments.
 For example, \cite{li2014deepreid,ahmed2015improved,varior2016gated,wu2017and,sun2017beyond} apply simple similarity metrics to discover local correspondences among  parts.  They assume misalignment within pre-defined strides or patches,  therefore these methods do not work well for larger misalignments. Next, attention methods \cite{li2018harmonious,zhao2017deeply,sun2017beyond,guo2018efficient} focus on learning salient regions, extracting strong activations in deep feature maps. The selected regions lack semantic interpretation, therefore in case of severe misalignment they may correspond to different body or object parts and should not be directly compared.
Methods from \cite{wei2017glad,su2017pose,zhao2017spindle,xuattention} rely on a body part detection, thus their performance depends on the accuracy of such detectors.
Despite this issue, we believe that body part locations can be used to align pedestrian samples in a robust way. Moreover, body joint estimators are being improved for other applications, which will also lead to better performance of re-ID methods relying on such detectors.
\newline
\textbf{GAN \cite{goodfellow2014generative}  in person re-ID. }    One of the first attempts to adopt GAN-generated samples for training CNN embedding in re-ID was \cite{zheng2017unlabeled}. Furthermore, \cite{liu2018pose} improves the generalization ability of the model by training with new generated poses. Other works, such as \cite{wei2018person,deng2018image} propose a  GAN model able to project pedestrian images between different dataset domains. Likewise, \cite{zhong2017camera} focuses on camera style.
\newline
 In contrast, to address partial re-ID problem, we propose to employ a body joint detector to align partial views and Cycle GAN model \cite{zhu2017unpaired} to hallucinate body parts missing in the partial view. 
\section{Partial matching network}
\begin{figure*}[h]
\begin{center}
{\includegraphics[width=11cm]{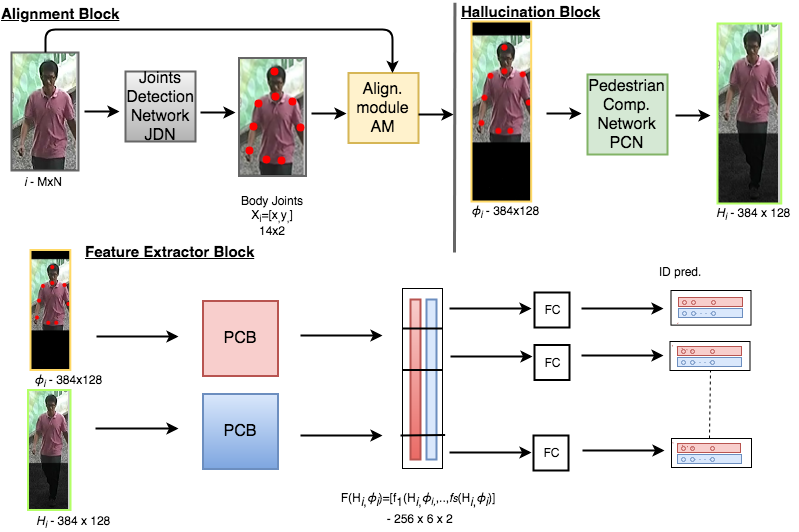}}
\end{center}
   \caption{Partial Matching Network.   \textit{Alignment Block (AB)}  aligns the input example $i$ of size $MxN$, \textit{Hallucinating Block (HB)} reconstructs missing parts and  \textit{Feature Extractor Block (FEB)} computes features from the aligned and reconstructed frames for several horizontal strides that are combined with fully connected layers.}
\label{fig:net}
\end{figure*}

In this section the proposed  \textit{Partial Matching Net (PMN)} is introduced with its three  components, i.e.  \textit{Alignment Block (AB)},  \textit{Hallucination Block (HB)} and \textit{Feature Extractor Block (FEB)} as illustrated in figure \ref{fig:net}. 
The alignment block \textit{AB} detects human body joints and aligns the input image to a reference frame such that locations of body parts in different images correspond, which facilitates matching. Missing areas of the aligned example in the reference frame are zero padded. The hallucination block then reconstructs the pedestrian appearance in the padded areas based on the information present in the frame and a learned model of a person.
Finally, a person representation is extracted from the reconstructed and aligned frames by \textit{FEB}. In the following, we present the three components in more detail.
\subsection{Alignment Block}\label{al_branch}
\begin{figure*}[h]
\centering
\begin{tabular}{cc}
{{\includegraphics[width=6.15cm]{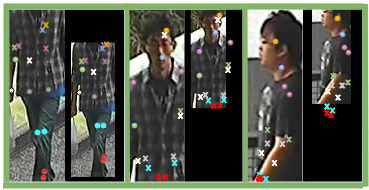}}}&
{{\includegraphics[width=6.15cm]{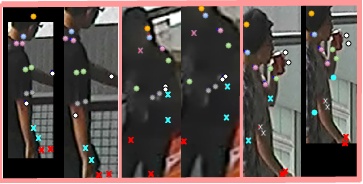}}}
\end{tabular}
\caption{Example pairs of partial views with detected and aligned joints: head (yellow), neck (sky blue), shoulders (pink), elbows (green), wrists (white), hips (grey), knees (turquoise) and ankles (red). (Left) Success cases. (Right) Failures.}
\label{fig:qualitative_evaluation}
\end{figure*}

Based on the results reported in the literature person re-ID is more reliable in images that are carefully cropped and person body parts are well aligned, in contrast to automatically detected bounding boxes with significant translation and scale change between different subjects or instances of the same ID.
We, therefore, attempt to perform the alignment automatically by employing state of the art \textit{Joints Detection Network (JDN)} \cite{wei2016convolutional} and then using the position of joints to align the input image in a reference frame. 
Specifically, JDN estimates positions $[x_k, y_k, m_k]$ of  $14$ body joints, corresponding to \{head, neck, rightshoulder, rightelbow, rightwritst, leftshoulder, leftebow, leftwrist, lefthip, leftknee, leftankle, righthip, rightknee, rightankle\}, where $x_k$ and $y_k$ are the joint coordinates, and $m_k \in [0,1]$ is a confidence value for part $k$.
From the training data, we estimate the average locations of body joints, which we use in a reference frame to which all examples can be aligned. Based on the confidence value of each coordinate $[x_k, y_k, m_k]$ for different joints, we select a subset of stable reference coordinates that are then used to estimate spatial transformation between the input frame and the reference frame.
Specifically, we assume confidence values of each body joint in the training set follow a normal distribution, and we consider a joint as reliable if included within $n=3$ standard deviations of the mean.
\newline
Regarding the spatial transformation for the alignment, as misalignment mostly derives from the inaccuracy of the detector and it is more significant along the y-direction, we assume it is a similarity transformation. In particular, parameters are estimated as follows: given the joints of the current sample $X_i$ and the average coordinates of joints $X_m$, the parameters $\Phi_i$ related to the similarity transformation are estimated by selecting the least square solution $\Phi_i=\underset{\Phi}{\arg\min}||\Phi X_i - X_m||^2$, i.e. minimizing the sum of squared residuals between the current and average coordinates.
\newline
Missing areas in the aligned frame are zero padded.
Since the transformation to align input frames is estimated from detected joints, the accuracy mainly depends on JDN. \textbf{Some qualitative results of alignments are shown in figure~\ref{fig:qualitative_evaluation}} with successfully detected joints in left and failure cases in right images. Specifically, in the first failure example, elbows are wrongly classified as hips, and as a result, the frame is zoomed in rather than zoomed out.
Note that JDN provides coordinates of body parts even if they are missing in the input image. We observe that the main misalignment in the input images is in the vertical direction, therefore the transformation is typically a vertical translation and a scale change. Furthermore,   the scale change is constrained to the same in vertical and horizontal direction to preserve the aspect ratio of the person.


\subsection{Hallucination block}
\begin{figure*}[h]
\begin{center}
{\includegraphics[height=4cm]{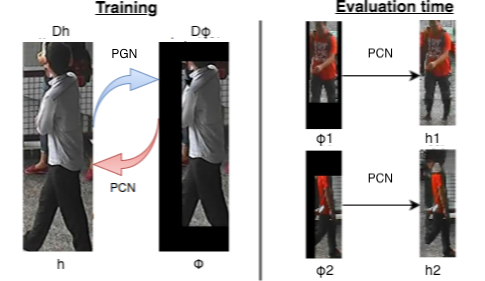}}
\end{center}
   \caption{PCN mapping from full body $h$ to partial observations $\phi$  and reverse mapping $PGN$.}
\label{fig:model Pedestrian completion network}
\end{figure*}

Aligned examples are much better suited for extracting and comparing features between corresponding body parts. However, partial views may cover different body parts, therefore the similarity scores may vary not only due to viewpoint change, illumination, and occlusions but also due to the fact that a different number of filters is deactivated by zero padding in different examples. One could address this issue by attempting to normalize the scores depending on what areas and parts are present. This, however, introduces more complexity to the re-ID process,  and requires specific solutions rather than  allowing to use existing re-ID approaches. Instead, we introduce a hallucination block \textit{HB} and propose to reconstruct the missing parts of the image. The advantage of our approach is that any state of the art person re-ID method can then be used. Moreover, given full views of pedestrians, it is straightforward to generate a large number of partial views for training a hallucination network.
%
\begin{figure*}[h]
\begin{center}
\begin{tabular}{c}
{{\includegraphics[height=2.15cm]{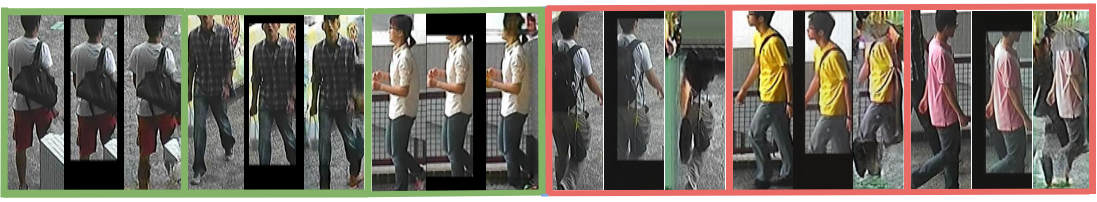}}}\\
{{\includegraphics[height=2.15cm]{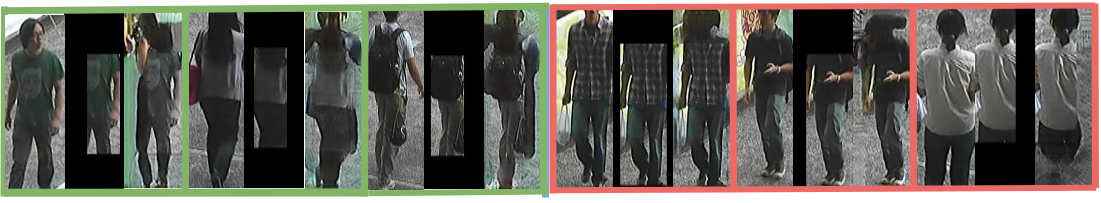}}}
\end{tabular}
\end{center}
   \caption{(Top) In green frame, examples of  successful frames, partial aligned views and reconstructed from 1/2 area missing. In red frame, poor reconstruction examples.
   (Bottom) In green frame, more examples of original frames with aligned 1/4 views and the reconstructed ones. In red frame, poor reconstruction examples. }
\label{fig:Pedestrian completion network}
\end{figure*}
\newline
Impressive results have been achieved in image inpainting and generating new images with GANs, e.g. by synthesizing new views of pedestrians  \cite{liu2018pose}.  Partial views often contain useful information that allows to extrapolate the appearance of a person and reconstruct missing parts in a similar way humans do.
 \textit{Cycle-Consistency Adversarial Network} \cite{zhu2017unpaired} was recently introduced for transferring image appearance into a different season, generating realistic views from paintings and photo enhancement. 
\newline
The network aims to learn a mapping function, GAN, from domain ${\Phi}$ to  ${H}$, such that the output images are indistinguishable from images belonging
to  ${H}$. Since there are infinite mapping functions between $\Phi$ and ${H}$, the optimization process suffers from mode collapse,  where the mapping tends to transform all the inputs to the same image, never converging. To address this issue, \textit{"cycle consistency"}  imposes that the transformed input image can be used to reconstruct the original version. In practice, this is implemented by introducing inverse mapping, i.e. another GAN, and then jointly training both models with a cycle consistency loss.
\newline
We adapt this approach by defining two mapping functions \textit{Pedestrian Completion Network (PCN) and \textit{Partial Generation Network (PGN)}}, which transform aligned partial observations $\phi$ into full body frames $h$ and vice-versa, as illustrated in figure \ref{fig:model Pedestrian completion network}.
In particular, our aim is to learn $PCN$ that hallucinates missing parts of the image, given partially aligned samples $\phi \in \Phi$ and full body samples $h \in H$. The model is optimized with the following objective:
\begin{equation}
\footnotesize{
\mathcal{L}_{tot}=\mathcal{L}_{GAN}(PGN,D_\phi,h,\phi)+\mathcal{L}_{GAN}(PCN,D_h,h,\phi)+\lambda\mathcal{L}_{cyc}(PGN,PCN)}
\end{equation}
where $\mathcal{L}_{GAN}$ are the adversarial losses, and $\mathcal{L}_{cyc}$ is the cycle consistency loss.
 Generators $PGN$ and $PCN$ aim to produce images that belong to domains $\Phi$ and $H$, respectively,  while the discriminators $D_\phi$ and $D_h$ try to distinguish real from fake samples in their respective domains. Finally, the cycle consistency loss is defined as
\begin{equation}
\footnotesize{
    {L}_{cyc}(PGN,PCN)=|PGN(PCN(\phi))-\phi|_1+|PCN(PGN(h))-h|_1}
\end{equation} that is the sum of $l_1$ norms between the reconstructed and the original samples. Note that, $PCN$ model can be applied independently, using partial aligned observation $\phi$ as an input, generating a complete frame $h$ with the missing parts hallucinated. Trained on a large number of examples the network learns to preserve the consistency of appearance between the partial views and the reconstructed parts.
Some qualitative results of hallucinated outputs are shown in figure \ref{fig:Pedestrian completion network}, in case of half and 3/4 of the original area missing.  The task is significantly more challenging when only 1/4 of the original area is preserved. In particular, the module is effective in reconstructing legs, head, and feet in examples shown in green frames, but it fails to reproduce fine details such as texture clothes in examples shown in red frames. The reason why some reconstructed regions are smooth and lack of fine details might depends on the overfitting of the model to the available regions.

\subsection{Feature extractor block}
After reconstructed missing parts of the aligned frame, both hallucinated and aligned frames are given as input to the feature extractor block (FEB).
Our FEB includes two separate feature extraction networks, trained independently with aligned and hallucinated samples while using the same architecture in both.
We experiment with two networks, i.e. ResNet50 baseline from \cite{zhong2017re}, and $PBC$ \cite{sun2017beyond}, that was recently designed for person re-ID.
Given an input image $i$, $PCB$ computes a 3D tensor of activations and then splits it into a number of column vectors by applying average pooling along the same horizontal strides. 
 The dimension of each feature component is reduced and  the resulting  stripes are concatenated.  
\newline
We propose to combine feature representations from the same stride in the hallucinated and the aligned frame using a fully connected layer. $F(h,\phi)=[f_1(h,\phi),...,f_s(h,\phi)]$ as illustrated in figure \ref{fig:net}.
The model is then optimized by minimizing the sum of cross entropy loss over each concatenated feature component, related to the same horizontal stride. 
During the evaluation, we adopt $F(h,\phi)$ as final feature vectors and euclidean distance as a comparison metric. 
\section{Experimental Results}
In this section, we first present the implementation details and the datasets used for experiments and then we discuss our analysis, demonstrating the effectiveness of the alignment and hallucination in partial re-identification and proving that our partial matching network is a valid solution.
Finally, we report the performance on standard benchmarks and compare our method against other re-ID systems for full \cite{zhong2017re,sun2017beyond} (without re-ranking) and partial views \cite{zheng2015partial,he2018deep} over several settings and datasets. 
\subsection{Implementation details} 
\textbf{PCN} network was trained with Adam optimizer and one partial observation per training sample with a learning rate of $0.0002$ and linear decay to zero over $100$ epochs. Other parameter settings can be found in \cite{sun2017beyond}.
\newline
\textbf{FEB} network were trained for $60$ epochs according to dataset protocols. The base learning rate was initialized at $0.1$ and kept the same for the first $40$ epochs then decayed to $0.01$. The model was pre-trained on ImageNet, and then, each $PCB$  was trained either with partial or hallucinated samples. The number of horizontal strides is fixed to $n=6$ and the final feature vector has  $256x6$ components. PCB networks for aligned and hallucinated frames were jointly fine-tuned for further $60$ epoch with a learning rate $0.005$, and decay by $0.01$ after $40$ epochs. The final feature vector $F(h,\phi)$ has dimensionality $256x6x2$.
\subsection{Datasets} 
\textbf{Partial REID} \cite{zheng2015partial} contains $600$ images of $60$ people taken from arbitrary viewpoints with different background/occlusions at a university campus. In particular, each person appears in $10$ frames, including $5$ partial observations and $5$ full-body images.
\newline
\textbf{Partial i-LIDS} \cite{zheng2015partial}  is a simulated dataset derived from i-LIDS, originally having $476$ pedestrian images of $119$ people acquired by different and non-overlapping cameras with a severe amount of occlusion. The simulated partial version, named \textit{Partial i-LIDS}, is achieved by generating as query images one partial observation for each pedestrian i.e. by selecting the most occluded view and manually cropping the not occluded part.
%
\begin{figure*}[h]
\begin{center}
{\includegraphics[width=12cm]{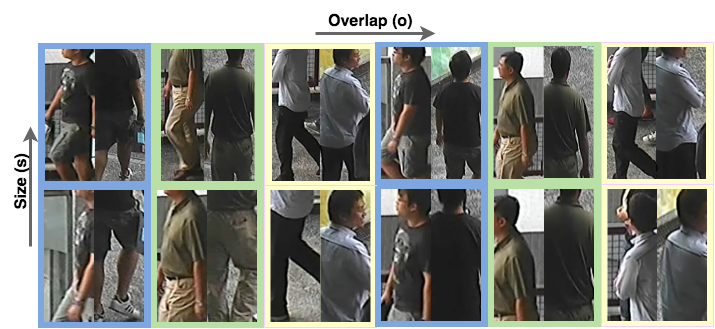}}
\end{center}
   \caption{Examples from CropCUHK03 datasets. (Bottom) 25\% of full body view, $s=0.25$, $o_{min}=[0,0.5]$.  (Top) 50\% of full body views, $s=0.5,o_{min}=[0.25,0.5]$. The overlap is increasing from left to right. Even for humans it is very challenging to match some  partial pedestrian images with little overlap between  two camera views.}
\label{fig:crop_cuhk03_examples}
\end{figure*}
\textbf{CropCUHK03} is our proposed synthetic dataset with partial crops from \textit{CUHK03} \cite{li2014deepreid} of different size and overlap between views.
Originally the dataset contains $14,097$ frames of $1467$ pedestrians, taken with $2$ different camera views and collected at CUHK campus. 
Several settings are generated maintaining the same number of individuals and frames as in CUHK03, which we refer to as  $CropCUHK03_{(s,o)}:s\in\{0.25,0.5\} \land o_{\min} \in\{0.0,0.5\}$, where $s$ is  the fraction of the area cropped from CUHK03 labeled frames i.e. the maximum possible overlap between camera views, and $o$ is the minimum overlap. Note that two crops from different views do not need to fully overlap. The crops are generated from random locations by keeping the aspect ratio the same as the original frame and making sure that the overlap between crops is at least $o$. This typically corresponds to less overlap between cropped parts as they come from different views.
 Figure \ref{fig:crop_cuhk03_examples} shows some examples from the generated partial views. The level of difficulty is controlled by parameters, e.g. for $s=0.25,o_{\min}=0.5$ the crop is $0.25$ of the full view and the overlap varies between $0.5$ and $1$ of the crop. The extreme cases are very difficult to match even for humans, given little overlap between camera views.
\subsection{Evaluation on CropCUHK03 datasets}
\textbf{Experimental setting.} 
We follow the new training/testing protocol \cite{zhong2017re}, where training and testing sets have $767$ and $700$ identities, respectively, and using our synthetic datasets characterized by crops of increasing difficulty.
We report results  for a popular re-ID baseline Resnet50 \cite{zhong2017re} without re-ranking, and a recent architecture PCB \cite{sun2017beyond}.
All images are resized to $256x256$ and $384x128$ as input to the Resnet50 and PCB, respectively.
\newline
\textbf{Evaluation of individual blocks.}
 We provide a systematic evaluation of the proposed solutions, particularly proving that alignment and hallucination are effective strategies for the task. 
 Our analysis shows to what extent the alignment, hallucination, and the proposal feature extractor block are effective solutions for the partial re-id task, where only a few of human body is available.
\newline
\textbf{Alignment.} As discussed in section~\ref{al_branch} the main error comes from inaccurate joint detection. To measure the extent to which this affects the overall results we compare our method, termed {\em align}, to the case of manual alignment, which is assumed as the one in labeled original frames. We also include a naive solution, denoted {\em baseline}, which simply re-scales the cropped views to the required input size. Table \ref{tab:alignment branch} shows the results for different feature extractors, i.e. ResNet50 and PCB, varying overlap of partial views, {\em baseline}, {\em man align} and {\em align}.
\newline
First important observation is that PCB significantly outperforms ResNet50 in all experiments. Note that PCB extracts a feature vector per stride while ResNet50 computes global representation. 
\newline
\textbf{Next, view alignment is particularly effective when the misalignment is large, that is the overlap between partial views is small} i.e. $o_{\min}\leq 0.25$, and brings a significant improvement regardless how much of the full frame is used. Resized {\em baseline} gives slightly better results for overlap at least $0.5$. This may be due to the filters that are deactivated by zero padding in the aligned views and general robustness of the networks to some misalignment.
\newline
Finally, alignment based on automatically detected joints by JDN~\cite{wei2016convolutional} is less effective, due to the errors introduced by JDN as discussed in section \ref{al_branch}. 
JDN was designed to detect a fixed number of parts, therefore it is forced to detect parts even when they are not present in the frames.
This however shows that despite impressive results from recent human pose estimation networks there is still a gap to bridge, before it brings clear benefits to partial person re-ID. 
In the following experiments, we focus on the cases of minor overlap i.e. $s\in\{0.5,0.25\}, o_{\min}\in\{0.25,0\}$ where the alignment brings significant improvement.
\begin{table}[]
\centering
\begin{tabular}{|ccccccccccccccc|}
\hline
\multicolumn{3}{|c|}{\textit{settings}}                         & \multicolumn{4}{c|}{\textit{baseline}}                                                                      & \multicolumn{4}{c|}{\textit{man align}}                                                                       & \multicolumn{4}{c|}{\textit{align}}              \\
\textit{mod} & \textit{s} & \multicolumn{1}{c|}{\textit{o\textsubscript{min}}} & {\textit{r1}}         & {\textit{r5}}         & {\textit{r10}}        & 
\multicolumn{1}{c|}{\textit{map}}        & {\textit{r1}}         & {\textit{r5}}         & {\textit{r10}}        & \multicolumn{1}{c|}{\textit{map}}        & \textit{r1} & \textit{r5} & \textit{r10} & \textit{map} \\ 
\hline
R50           & .5         & \multicolumn{1}{c|}{.5}         & {\textbf{11.8}} & 22.9                & {\textbf{31.2}} & \multicolumn{1}{c|}{{\textbf{10.7}}} & 11.5                & {\textbf{23.1}} & 29.8                & \multicolumn{1}{c|}{10.3}                & 8.2         & 17.8        & 23.6         & 7.1          \\
R50           & .25        & \multicolumn{1}{c|}{.5}         & {\textbf{6.6}}  & {\textbf{16.1}} & {\textbf{22.4}} & \multicolumn{1}{c|}{{\textbf{6.5}}}  & 6.1                 & 14.9                & 19.8                & \multicolumn{1}{c|}{6.1}                 & 3.3         & 8.8         & 12.9         & 2.6          \\ \hline
R50           & .5         & \multicolumn{1}{c|}{.25}         & 9.6                 & 19.7                & 27.8                & \multicolumn{1}{c|}{9.3}                 & { \textbf{11.2}} & {\textbf{22.1}} & { \textbf{29}}   & \multicolumn{1}{c|}{{  \textbf{10.4}}} &       7.3     &   15.4          &   21.8           &    6.1          \\
R50           & .25        & \multicolumn{1}{c|}{0}          & 5.9                 & 14.2                & 20.3                & \multicolumn{1}{c|}{5.2}                 & {  \textbf{6}}    & {  \textbf{14.7}} & {  \textbf{20.3}} & \multicolumn{1}{c|}{{  \textbf{5.7}}}  &     1.9        &   6.2          &      8.9        &       1.9       \\ \hline
PCB           & .5         & \multicolumn{1}{c|}{.5}         & {  \textbf{31.6}} & {  \textbf{55.8}} & {  \textbf{65.5}} & \multicolumn{1}{c|}{{  \textbf{21}}}   & 28.7                & 52                  & 62.8                & \multicolumn{1}{c|}{19.4}                & 26.1        & 47.6        & 57.4         & 16.9           \\
PCB           & .25        & \multicolumn{1}{c|}{.5}         & {  \textbf{23.7}} & {  \textbf{45.7}} & {  \textbf{55.8}} & \multicolumn{1}{c|}{{  \textbf{16.7}}} & 15.4                & 33.9                & 43.7                & \multicolumn{1}{c|}{10.2}                & 10.8           & 24.7        & 33.8         & 6.9          \\ \hline
PCB           & .5         & \multicolumn{1}{c|}{.25}         & 25.9                & 49.1                & 60.3                & \multicolumn{1}{c|}{16.5}                & {  \textbf{28.7}} & {  \textbf{52.4}} & {  \textbf{62.7}} & \multicolumn{1}{c|}{{  \textbf{19.3}}} &     22.7        &           43.4  &      53.5        &    25   \\
PCB           & .25        & \multicolumn{1}{c|}{0}          & 11.8                & 27.2                & 36.2                & \multicolumn{1}{c|}{7.2}                 & {  \textbf{12.3}} & {  \textbf{27.3}} & {  \textbf{36.3}} & \multicolumn{1}{c|}{{  \textbf{7.9}}}  &    7.7       & 18.4            &  25.1           & 4.5 \\
\hline
\end{tabular}

 \caption{Partial re-ID with alignment. The $baseline$ corresponds to partial observations resized to a fixed size,  manual alignment $man align$ assumes perfectly detected body joints and alignment ($align$) uses JDN (joint detection network). PCB significantly outperforms ResNet50 (R50), and the benefits of alignment are visible when the overlap between views is small i.e. $o_{\min}<0.25$. JDN introduces errors that affect the overall performance.}
\label{tab:alignment branch}
\end{table}
\begin{table}[h]
\centering
\footnotesize
\label{my-label}
\begin{tabular}{|c|cccc|cccc|}
\hline
{Methods}    & \multicolumn{4}{c|}{CropCUHK03 .5,.25}                        & \multicolumn{4}{c|}{CropCUHK03 .25,0}                        \\
                            & r1            & r5            & r10           & map           & r1            & r5            & r10           & map          \\ \cline{2-9} 
Res50+Eucl \cite{zhong2017re}                  & 9.6           & 19.7          & 27.8          & 9.3           & 5.9           & 14.2          & 20.3          & 5.2          \\
Res50+Eucl+re-rank \cite{zhong2017re}          & 11.6          & 19            & 24.4          & 12.8          & 6.8           & 12.9          & 18.4          & 7.2          \\
Res50+XQDA \cite{zhong2017re}                  & 16.6          & 30.7          & 38.4          & 14.5          & 8.9           & 19.7          & 26.4          & 7.4          \\
Res50+XQDA+re-rank \cite{zhong2017re}          & 18.6          & 30            & 37.1          & 20            & 10.2          & 20            & 25.7          & 10.4         \\ \hline
PCB baseline \cite{sun2017beyond}                & 25.9          & 49.1          & 60.3          & 16.5          & 11.8          & 27.2          & 36.2          & 7.2          \\
PCB align.                & 20.1          & 40.2          & 50            & 13            & 6.4           & 16            & 22.5          & 3.7          \\
PCB man align                 & 28.7          & 52.4          & 62.7          & 19.3          & 12.3          & 27.3          & 36            & 37.9         \\
PCB align+hall           & 15.4          & 32.3          & 41.3          & 9.3           & 3.2           & 9.7           & 14.8          & 1.7          \\
PCB man align+hall            & 25            & 47.6          & 58.1          & 15.9          & 6.8           & 17.9          & 24.8          & 4            \\ \hline
Our align+hall         & 22            & 42.5          & 52.5          & 14.4          & 5.4           & 14.5          & 20.6          & 3.1          \\
\textbf{Our man align+hall} & \textbf{31.3} & \textbf{55.6} & \textbf{65.4} & \textbf{21.2} & \textbf{13.2} & \textbf{28.4} & \textbf{37.5} & \textbf{8.4} \\ \hline
\end{tabular}
\caption{Partial re-ID with alignment and hallucination. Comparison to baseline as well as different variants of ResNet50 and our proposed approach. Combined alignment and hallucination consistently improves the results.}
\label{t:comparison}
\end{table}\newline
\textbf{Hallucination.}
Training and testing directly the baseline with hallucinated samples leads to a drop in re-ID rates, probably due to overfitting to the hallucinated samples.
Nevertheless, when FEB is applied to hallucinated and aligned samples (Our align + hall), it brings an improvement by nearly $3\%$- compared to the alignment block only (PCB align) and by $6\%$ compared to the baseline (PCB baseline), \textbf{proving that hallucinated samples along with our combining strategy (FEB) are effective for re-identification task.} As shown, features coming from hallucination block need to be carefully selected through our combining architecture which specifically emphasize most salient features, with a further fully connected layer concatenating feature vectors coming from both the hallucination and alignment branches. By following such a procedure, hallucinated features components, related to false cues and introducing ambiguity, are filtered, and only the most effective are emphasized.

\noindent\textbf{Comparative evaluation.}
Our methods are compared with different re-ID approaches. The results in table \ref{t:comparison} confirm that the PCB based models outperform ResNet50, even when combined with learned projections XQDA, and re-ranking strategies.
\textbf{Our architecture, jointly combining features from the aligned and hallucinated frames, significantly boosts the performance}, e.g. Our man align+hall improves rank 1 of PCB baseline from $25.9\%$ to $31.3\%$.
As expected, the results for 1/4 view and low overlap between views $s=0.25, o_{min}=0$ are significantly lower than for $s=0.5, o_{min}=0.25$ but our approach still improves upon the other methods. Specifically $rank 1$ increases by $1.5\%$ compared to PCB baseline.

\subsection{Evaluation on partial ReID and i-LIDS}
We also include a comparison on Partial ILIDS and Partial ReID datasets \cite{zheng2015partial}.
\newline
\textbf{Partial ReID.} 
All our analysis are shown in single-shot setting, in which the gallery set includes only one image for each person. 
In table \ref{tab:Evaluation on existing partial re-id datasets}  we compare our method against other partial re-ID methods i.e. SVM \cite{zheng2015partial} , AMC \cite{zheng2015partial}, and DSR \cite{he2018deep}, which achieve scores of 24.3\%, 33.3\% and 43\%, respectively. Results show that our approach,  assuming correctly detected body joints (PCB man align), gives significantly better score of $63\%$. There is still an improvement of 7\% when real joint detector is used.
\begin{table}[h]
\footnotesize
\centering
\label{my-label}
\begin{tabular}{|l|llll|llll|}
\hline
\multicolumn{1}{|c|}{{Method}} & \multicolumn{4}{l|}{Partial ReID}                                                   & \multicolumn{4}{l|}{Partial iLIDS}                                                  \\ \cline{2-9} 
\multicolumn{1}{|c|}{}                        & \multicolumn{1}{l|}{r1} & \multicolumn{1}{l|}{r5} & \multicolumn{1}{l|}{r10} & map  & \multicolumn{1}{l|}{r1} & \multicolumn{1}{l|}{r5} & \multicolumn{1}{l|}{r10} & map  \\ \hline
SWM \cite{zheng2015partial}                                             & 24.3                    & 52.3                    & 61.3                     & -  & 33.6                    & 53.8                    & 63                       & -   \\
AMC \cite{zheng2015partial}                                        & 33.3                    & 52                      & 62                       & -      & 46.8                    & 69.6                    & 81.8                     & -   \\
AMC+SWM \cite{zheng2015partial}                                    & 36                      & 60                      & 70.7                     & -      & 49.6                    & 72.3                    & 70.7                     & -   \\
DSR (single-scale) \cite{he2018deep}                       & 39.3                      & 65.7                    & 76.7                     & -    & 51.1                    & 70.7                    & 82.4                     & -     \\
DSR (multiple-scale) \cite{he2018deep}                      & 43                      & 75                      & 76.7                     & -    &  \textbf{54.6}                    &  \textbf{73.1}                    &  \textbf{85.7}                     & -     \\ \hline

PCB baseline \cite{sun2017beyond}                               &  54.9                    & 85.3                    & 93.2                     & 57.2& 30.6                    & 56.2                    & 68.4                     & 33.6 \\

PCB man align \cite{sun2017beyond}                                 & 61.7                    & 89.7                    & 95.9                     & 63.8 & 35.5                    & 61.7                    & 74.9                     & 37.4 \\

PCB+align.                               &         49.6                &  82.3                       &                92.2          &     52.2     &      35.3                   &      62                   & 73.6                         &   37  \\ 
PCB+align + hall                                    &             44.6                     &                 76.1         &    88.7 &  47.6         &  33.8                       &    59.4                     &  72.3                        & 35.2           \\  \hline
Our (man align +hall)                            &       \textbf{63}&          \textbf{89.1}               &     \textbf{95.3}                      &  \textbf{64.7}  &  35.3                        &     62.7                &     73.9                    &    37.4       \\
Our (align +hall)                                                           &   50                      &  82.3 &    92.2&    52.7     &        38.4                   &     64.5                     &  76.1    &          39.3    \\ \hline
\end{tabular}
\caption{Evaluation on  Partial ReID  and i-LIDS datasets. Our method performs best on Partial ReID but not on iLIDS due to low quality views and small training set.}
\label{tab:Evaluation on existing partial re-id datasets}
\end{table}
\noindent\textbf{Partial i-LIDS} has lower quality images than other datesets, which is reflected in the performance of all compared methods.  This dataset significantly differs from CUHK03, which we have adopted for pre-training the PCB. Furthermore, the number of samples of this dataset is too low to fine-tune our CNNs.
In contrast, DSR implements a network of $13$  layers only i.e. fewer parameters, and training with small datasets such as Partial i-LIDS is more effective. 
\section{Conclusion}
We have proposed an approach to partial person re-ID by combining body pose detection, spatial alignment, hallucination and feature extraction based on neural networks.
We have extensively analyzed the impact of misalignment and hallucination in partial Re-id task, proving that view alignment is particularly effective in case image persons are severely misaligned regardless of how much content is missing from the partial observations. Furthermore, our proposed block (FEB) jointly combining features from the aligned and hallucinated frames significantly improve the performance.
Finally, we have evaluated different variants of the approach and compared to state of the arts on three different datasets. 
The overall results demonstrate that hallucinating missing content of partial observation is an effective strategy for partial person re-identification.
We believe, however, there is still room for improving the proposed PCN implementation, for example, a loss function could be designed to drive the Cycle-Consistency Adversarial Network to mostly attend missing regions of the frame, avoid local minima and to enhance the network ability to reconstruct fine details. 
Since we have observed the accuracy of JDN on cropped person images is limited, we hope this work would also encourage research community to extend human body joint detector to effectively work on such challenging cases.

\bibliographystyle{splncs}

\end{document}